%% file: main.tex
\documentclass[lang=en,11pt]{elegantbook}
\usepackage{caption}
\usepackage{graphicx}

\input{math_commands}

\title{ Mathematical Cookbook for \\Snapshot Compressive Imaging}
\subtitle{Mathematics in Snapshot Compressive Imaging (SCI)}

\author{Yaping Zhao}
\date{Mar 19, 2023}
\version{0.50}
\bioinfo{Bio}{Yaping Zhao is a life-long student of physics, and her career encompasses engineering, thinking and writing.\\When she's not busy puzzling over academic research, Yaping reads interesting books.}

\extrainfo{What I cannot create, I do not understand. Know how to solve every problem that has been solved. --- Richard Feynman }

\cover{cover.jpg}

\begin{document}

\maketitle

\frontmatter
\tableofcontents

\mainmatter
\input{sections/intro}
\input{sections/optimization}
\input{sections/deq}

\newpage

{
% \small
% \bibliographystyle{ieee_fullname}
\bibliography{main}
}

\end{document}

%% file: math_commands.tex
\newcommand{\ie}{{\em i.e.}}

\newcommand{\Amat}{{\bf A}}

\newcommand{\Imat}{{\bf I}}

\newcommand{\Mmat}[0]{{{\bf M}}}

\newcommand{\av}{\boldsymbol{a}}

\newcommand{\ev}[0]{{\boldsymbol{e}}}

\newcommand{\rv}[0]{{\boldsymbol{r}}}

\newcommand{\uv}[0]{{\boldsymbol{u}}}
\newcommand{\vv}{\boldsymbol{v}}
\newcommand{\wv}{\boldsymbol{w}}

\newcommand{\xv}{\boldsymbol{x}}
\newcommand{\yv}{\boldsymbol{y}}

\newcommand{\Phimat}{\boldsymbol{\Phi}}
\newcommand{\Psimat}{\boldsymbol{\Psi}}

\newcommand{\alphav}{\boldsymbol{\alpha}}

\DeclareMathOperator*{\argmin}{arg\,min}

%% file: sections/intro.tex
\chapter{Introduction}

\section{Preface}
The author intends to provide you with a beautiful, elegant, user-friendly cookbook for mathematics in Snapshot Compressive Imaging (SCI). Currently, the cookbook is composed of \textbf{introduction, conventional optimization, and deep equilibrium models}. The latest releases are strongly recommended! For any other questions, suggestions, or comments, feel free to email the author.
% The author intends to provide you with a beautiful, elegant, user-friendly cookbook for mathematics in Snapshot Compressive Imaging (SCI). Currently, the cookbook is composed of \textbf{conventional optimization, plug-and-play framework, deep equilibrium models}, using regularization-based optimization algorithms, plug-and-play (PnP) framework, and deep equilibrium (DEQ) models for SCI respectively. The latest releases are strongly recommended! For any other questions, suggestions, or comments, feel free to email me.

Email: \email{zhaoyp18@tsinghua.org.cn}

Because the author is too lazy to write a tediously long introduction, the author assume that readers of this book own preliminary knowledge about the mathematical model of SCI. If you do not, the author highly recommend you to read the introduction session of the paper~\cite{zhao2022deep} or any other SCI publications before you begin this mathematical journey.

%% file: sections/optimization.tex
\chapter{Conventional Optimization}
To begin with, imaging we observe a corrupted set of measurements $\yv$ of an image $\xv$ under a linear measurement operator $\Phimat$ with some noise $\ev$ according to

\begin{align}
    \mathbf{y} = \mathbf{\Phi} \mathbf{x} + \mathbf{e},
\end{align}

Suppose we have a known regularization function $R$ that could be applied to an image x. Then we could compute an image estimate $\hat x$ by solving the optimization problem

\begin{align}
\label{eq:opt}
    \mathbf{\hat{x}} = \mathop{\arg\min}_{ \mathbf{x} } \frac{1}{2}|| \mathbf{y} - \mathbf{\Phi} \mathbf{x} ||_2^2 + R(\mathbf{x}).
\end{align}

%--------Gradient Descent------------------------------------------------------------------------
\section{Gradient Descent}
If $R$ is differentiable, this can be accomplished via gradient descent. We could get

\begin{align}
\begin{aligned}
    &\frac{\partial}{\partial \xv} [\frac{1}{2}|| \yv - \Phimat \xv ||_2^2 + R(\xv)] \\
    =&-\frac{1}{2}\cdot2\cdot\Phimat^T(\yv - \Phimat \xv) + \frac{\partial{R(\xv)}}{\partial \xv} \\
    =&-\Phimat^T(\yv - \Phimat \xv) + \bigtriangledown R(\xv).
\end{aligned}
\end{align}

That is, we start with an initial estimate $\xv^{(0)}$ such as $\xv^{(0)}=\Phimat^T \yv$ and choose a step size $\alpha > 0$, such that for iteration $k=1,2,3...$, we set

\begin{align}
\begin{aligned}
    \xv^{(k+1)} &= \xv^{(k)} - \alpha[-\Phimat^T(\yv - \Phimat \xv) + \bigtriangledown R(\xv)] \\
    &=\xv^{(k)} + \alpha\Phimat^T(\yv - \Phimat \xv)  - \alpha \bigtriangledown R(\xv).
\end{aligned}
\end{align}

%--------ADMM--------------------------------------------------------------------------------------
\section{ADMM: Alternating Directions Method of Multipliers}
By introducing an auxiliary parameter $\vv$, the unconstrained optimization in Eq.\ref{eq:opt} can be converted into

\begin{align}
\begin{aligned}
    \label{eq:opt2}
    (\hat{\xv}, \hat{\vv}) = \mathop{\arg\min}_{\xv, \vv} \frac{1}{2}|| \yv - \Phimat \xv ||_2^2 + R(\vv), \\
    s.t.\ \xv=\vv.
\end{aligned}
\end{align}

\subsection{Augmented Lagrangian Method}
Using augmented Lagrangian method, we introduce an parameter $\uv$ to be updated and another $\rho$ to be manually set, and get the augmented Lagrangian function of Eq.\ref{eq:opt2} as

\begin{align}
\label{eq:lag}
    L_\rho(\xv, \vv, \uv) &= \frac{1}{2}|| \yv - \Phimat \xv ||_2^2 + R(\vv) + \uv^T(\xv-\vv) + \frac{\rho}{2}||\xv-\vv||_2^2
\end{align}

To facilitate updates of $\xv$, we rewrite Eq.\ref{eq:lag} as

\begin{align}
\begin{aligned}
\label{eq:lag1}
    L_\rho(\xv, \vv, \uv) &= \frac{1}{2}|| \yv - \Phimat \xv ||_2^2 + R(\vv) + \uv^T\xv - \uv^T\vv  + \frac{\rho}{2}(\xv^T\xv - 2\cdot \xv^T\vv + \vv^T\vv) \\
    &= \frac{1}{2}|| \yv - \Phimat \xv ||_2^2 + R(\vv) + \frac{\rho}{2}[\xv^T\xv - 2(\vv- \frac{1}{\rho}\uv)\xv + (\vv - \frac{1}{\rho}\uv)^2 - (\vv - \frac{1}{\rho}\uv)^2] - \uv^T\vv \\
    &= \frac{1}{2}|| \yv - \Phimat \xv ||_2^2 + R(\vv) + \frac{\rho}{2}||\xv - (\vv - \frac{1}{\rho}\uv) ||_2^2 -\frac{\rho}{2}(\vv - \frac{1}{\rho}\uv)^2 - \uv^T\vv.
\end{aligned}
\end{align}

Similarly, to facilitate updates of $\vv$, we have 

\begin{align}
\begin{aligned}
\label{eq:lag2}
    L_\rho(\xv, \vv, \uv) &= \frac{1}{2}|| \yv - \Phimat \xv ||_2^2 + R(\vv) + \uv^T\xv - \uv^T\vv  + \frac{\rho}{2}(\vv^T\vv - 2\cdot \xv^T\vv + \xv^T\xv) \\
    &= \frac{1}{2}|| \yv - \Phimat \xv ||_2^2 + R(\vv) + \frac{\rho}{2}[\vv^T\vv - 2(\xv+ \frac{1}{\rho}\uv)\vv + (\xv + \frac{1}{\rho}\uv)^2 - (\xv + \frac{1}{\rho}\uv)^2] + \uv^T\xv \\
    &= \frac{1}{2}|| \yv - \Phi \xv ||_2^2 + R(\vv) + \frac{\rho}{2}||\vv - (\xv + \frac{1}{\rho}\uv) ||_2^2 -\frac{\rho}{2}(\xv + \frac{1}{\rho}\uv)^2 + \uv^T\xv.
\end{aligned}
\end{align}

Then ADMM solves it by the following sequence of sub-problems:

\begin{align}
\begin{aligned}
    \xv^{(k+1)} &= \mathop{\arg\min}_{\xv} L_\rho(\xv, \vv^k, \uv^k) \\
    &= \mathop{\arg\min}_{\xv} \frac{1}{2}|| \yv - \Phimat \xv ||_2^2 + R(\vv^k) + \frac{\rho}{2}||\xv - (\vv^k - \frac{1}{\rho}\uv^k) ||_2^2 -\frac{\rho}{2}(\vv^k - \frac{1}{\rho}\uv^k)^2 - {\uv^k}^T\vv^k \quad(Eq.\ref{eq:lag1})\\
    &= \mathop{\arg\min}_{\xv} \frac{1}{2}|| \yv - \Phimat \xv ||_2^2 + \frac{\rho}{2}||\xv - (\vv^k - \frac{1}{\rho}\uv^k) ||_2^2 \quad(delete\ terms\ not\ related\ to\ \xv),
\end{aligned}
\end{align}

\begin{align}
\begin{aligned}
    \vv^{(k+1)} &= \mathop{\arg\min}_{\vv} L_\rho(\xv^k, \vv, \uv^k) \\
    &= \mathop{\arg\min}_{\vv} \frac{1}{2}|| \yv - \Phimat \xv ||_2^2 + R(\vv) + \frac{\rho}{2}||\vv - (\xv^k + \frac{1}{\rho}\uv^k) ||_2^2 -\frac{\rho}{2}(\xv^k + \frac{1}{\rho}\uv^k)^2 + {\uv^k}^T\xv \quad(Eq.\ref{eq:lag2})\\
    &= \mathop{\arg\min}_{\vv} R(\vv) + \frac{\rho}{2}||\vv - (\xv^k + \frac{1}{\rho}\uv^k) ||_2^2 \quad(delete\ terms\ not\ related\ to\ \vv),
\end{aligned}
\end{align}

\begin{align}
\begin{aligned}
    \uv^{(k+1)} &= \mathop{\arg\min}_{\uv} L_\rho(\xv^k, \vv^k, \uv) \\
    &= \uv^k + \alpha\frac{\partial}{\partial \uv} L_\rho(\xv^{(k+1)}, \vv^{(k+1)}, \uv) \quad(gradient\ descent) \\
    &= \uv^k + \alpha \frac{\partial}{\partial \uv} (\frac{1}{2}|| \yv - \Phimat \xv^{(k+1)} ||_2^2 + R(\vv^{(k+1)}) + \uv^T(\xv^{(k+1)}-\vv^{(k+1)}) + \frac{\rho}{2}||\xv^{(k+1)}-\vv^{(k+1)}||_2^2) \\
    &= \uv^k + \alpha(\xv^{(k+1)}-\vv^{(k+1)}) \\
    &= \uv^k + \rho(\xv^{(k+1)}-\vv^{(k+1)}) \quad(set\ \alpha = \rho).
\end{aligned}
\end{align}

\subsection{Scaled Form}
ADMM can be written in a slightly different form, which is often more convenient, by combining the linear and quadratic terms in the augmented Lagrangian and scaling the dual variable. 
Defining the residual $\rv = \xv-\vv$ and recall Eq.\ref{eq:lag}, we have

\begin{align}
\begin{aligned}
    \uv^T(\xv-\vv) + \frac{\rho}{2}||\xv-\vv||_2^2 &= \uv^T\rv + \frac{\rho}{2}||\rv||_2^2 \\
    &=\frac{\rho}{2}\rv^T\rv + \uv^T\rv \\
    &=\frac{\rho}{2}(\rv^T\rv + 2\cdot\frac{1}{\rho}\uv^T\rv + \frac{1}{\rho^2}\uv^T\uv-\frac{1}{\rho^2}\uv^T\uv) \\
    &=\frac{\rho}{2}(\rv^T\rv + 2\cdot\frac{1}{\rho}\uv^T\rv + \frac{1}{\rho}\uv^T\uv) - \frac{1}{2\rho}\uv^T\uv \\
    &=\frac{\rho}{2}||\rv+\frac{1}{\rho}\uv||_2^2 - \frac{1}{2\rho}||\uv||_2^2 \\
    &=\frac{\rho}{2}||\rv+\wv||_2^2 - \frac{\rho}{2}||\wv||_2^2 \quad(\wv = \frac{1}{\rho}\uv),
\end{aligned}
\end{align}

where $\wv = \frac{1}{\rho}\uv$ is the scaled dual variable. Using the scaled dual variable, we can rewrite Eq.\ref{eq:lag} as

\begin{align}
    L_\rho(\xv, \vv, \wv) &= \frac{1}{2}|| \yv - \Phimat \xv ||_2^2 + R(\vv) + \frac{\rho}{2}||\xv-\vv+\wv||_2^2 - \frac{\rho}{2}||\wv||_2^2
\end{align}

Then we can express ADMM as

% x^{(k+1)} = \mathop{\arg\min}_{x} L_\rho(x, v^k, w^k) \\
    % v^{(k+1)} = \mathop{\arg\min}_{v} L_\rho(x^k, v, w^k)
\begin{align}
\begin{aligned}
    \xv^{(k+1)} &= \mathop{\arg\min}_{\xv} L_\rho(\xv, \vv^k, \wv^k) \\
    &=\mathop{\arg\min}_{\xv} \frac{1}{2}|| \yv - \Phimat \xv ||_2^2 + R(\vv) + \frac{\rho}{2}||\xv-\vv+\wv||_2^2 - \frac{\rho}{2}||\wv||_2^2 \\
    &=\mathop{\arg\min}_{\xv} \frac{1}{2}|| \yv - \Phimat \xv ||_2^2 + \frac{\rho}{2}||\xv-\vv+\wv||_2^2 \quad(delete\ terms\ not\ related\ to\ \xv),
\end{aligned}
\end{align}
    
\begin{align}
\begin{aligned}
    \vv^{(k+1)} &= \mathop{\arg\min}_{\vv} L_\rho(\xv^k, \vv, \wv^k) \\
    &=\mathop{\arg\min}_{\vv} \frac{1}{2}|| \yv - \Phimat \xv ||_2^2 + R(\vv) + \frac{\rho}{2}||\xv-\vv+\wv||_2^2 - \frac{\rho}{2}||\wv||_2^2 \\
    &=\mathop{\arg\min}_{\vv} R(\vv) + \frac{\rho}{2}||\xv-\vv+\wv||_2^2 \quad(delete\ terms\ not\ related\ to\ \vv),
\end{aligned}
\end{align}

\begin{align}
\begin{aligned}
    \wv^{(k+1)} &= \wv^k + \alpha\frac{\partial}{\partial \wv} L_\rho(\xv^{(k+1)}, \vv^{(k+1)}, \wv) \quad(gradient\ descent) \\
    &= \wv^k + \alpha[\rho(\xv^{(k+1)}-\vv^{(k+1)}+\wv)-\rho \wv] \\
    &= \wv^k + \alpha\rho(\xv^{(k+1)}-\vv^{(k+1)}) \\
    &= \wv^k + \xv^{(k+1)}-\vv^{(k+1)} \quad(set\ \alpha\rho = 1).
\end{aligned}
\end{align}
%--------GAP----------------------------------------------------------------------------------
\section{GAP: Generalized Alternating
Projection}
Generalized alternating projection (GAP) can be recognized as a special case of ADMM, which works as a lower computational workload algorithm. Recall Eq. \ref{eq:opt2}, GAP updates $\xv^k$ and $\vv^k$ as follows:
\begin{itemize}
    \item Updating $\xv: \xv^k$ is updated via a Euclidean projection of $\vv^k$ on the linear manifold $\Mmat : \yv = \Phimat \xv$. That is,
\end{itemize}

\begin{align}
\label{eq:gap-x}
    \xv^{(k+1)} = \vv^k + {\Phimat}^\top (\Phimat {\Phimat}^\top)^{-1}(\yv-\Phimat \vv^k).
\end{align}

\begin{itemize}
    \item Updating $\vv$: After the projection, the goal of the next step is to bring $\xv^{(k+1)}$ closer to the desired signal domain. This could be achieved by employing an appropriate trained denoiser $\mathcal{D}_{k+1}$ and letting
\end{itemize}

\begin{align}
\label{eq:gap-v}
    \vv^{(k+1)} = \mathcal{D}_{k+1} (\xv^{(k+1)}).
\end{align}

Derivation of Eq. \ref{eq:gap-x}: given $\vv^k$, how to obtain the Euclidean projection of $\vv^k$ on the linear manifold? Since the Euclidean projection is essentially finding the shortest distance between $\Mmat : \yv = \Phimat \xv$ and $\vv$, this problem can be modeled as

\begin{align}
    \begin{aligned}
    \label{eq:gap-min}
        \mathop{\arg\min}_{\xv} \frac{1}{2}|| \xv - \vv ||_2^2,& \quad(minimize\ the\ distance\ between\ \xv\ and\ \vv) \\ 
    s.t.\ \Phimat \xv=\yv.& \quad(ensure\ finding\ the\ point\ \xv\ on\ \Mmat\ nearest\ to\ \vv)
    \end{aligned}
\end{align}

Easily we could get the Lagrangian function of Eq. \ref{eq:gap-min} by introducing an parameters $\lambda$,

\begin{align}
\label{eq:gap-lag}
    L(\xv, \lambda) = \frac{1}{2}|| \xv - \vv ||_2^2 + \lambda (\Phimat \xv-\yv).
\end{align}

Then the optimal conditions of Eq. \ref{eq:gap-lag} are:

\begin{align}
\label{eq:gap-01}
    \frac{\partial}{\partial \xv} L(\xv, \lambda)=(\xv-\vv) + \Phimat^\top\lambda=0, \\
    \label{eq:gap-02}
    \frac{\partial}{\partial \lambda} L(\xv, \lambda)=\Phimat \xv-\yv=0.
\end{align}

According to Eq. \ref{eq:gap-01}, we have $\xv = \vv-\Phimat^\top\lambda$, combined with $\Phimat \xv=\yv$, we could get
\begin{align}
    \begin{aligned}
    &\Phimat (\vv-\Phimat^\top\lambda) = \yv \\
    \Rightarrow\quad & \Phimat \vv - \Phimat \Phimat^\top\lambda = \yv \\
    \Rightarrow\quad & - \Phimat \Phimat^\top\lambda = \yv - \Phimat \vv \\
    \Rightarrow\quad & \lambda = - (\Phimat \Phimat^\top)^{-1}(\yv - \Phimat \vv).
    \end{aligned}
\end{align}

Combining $\lambda = - (\Phimat \Phimat^\top)^{-1}(\yv- \Phimat \vv)$ with $\xv = \vv-\Phimat^\top\lambda$, we could get

\begin{align}
    \xv^{(k+1)} = \vv^k + \Phimat^\top(\Phimat \Phimat^\top)^{-1}(\yv - \Phimat \vv^k).
\end{align}

%-------PGD-----------------------------------------------------------------------------------
% \section{Proximal Gradient Descent}
% To appear.

%% file: sections/deq.tex
\chapter{Deep Equilibrium Models}

\section{Review of Deep Learning based Algorithms}
Given the masks and measurements, plenty of algorithms including conventional optimization\cite{liu2018rank,Yang14GMMonline,Yang14GMM,yuan2016generalized}, end-to-end deep learning~\cite{qiao2020deep,zheng2021super, wang2022spatial, cheng2022recurrent, meng2021perception}, deep unfolding~\cite{meng2020gap,wu2021dense} and plug-and-play~\cite{Yuan2020_CVPR_PnP,PnP_SCI_TPAMI21, wu2022adaptive, yang2022revisit} are proposed for reconstruction.

To accommodate the state-of-the-art SCI architectures and to enable \textbf{low-memory stable} reconstruction, this chapter sets about utilizing deep equilibrium models (DEQ)~\cite{bai2019deep} for solving the inverse problem of video SCI. Specifically, we applied DEQ to two existing models for video SCI reconstruction: recurrent neural networks (RNN) and plug-and-play framework (PnP).

Given measurement $\yv\in {\mathbb R}^{n}$ with compression rate $B$ and sensing matrix $\Phimat\in {\mathbb R}^{n\times nB}$ as input, we consider an optimization iteration or neural network as:
\begin{align}
    \xv^{(k+1)} = f_\theta (\xv^{(k)}; \yv, \Phimat),\quad k = 0,1,\dots, \infty, 
\end{align}
where $\theta$ denotes the weights of embedded neural networks; $\xv^{(k)}\in {\mathbb R}^{nB}$ is the output of the $k^{th}$ iterative step or hidden layer, and $\xv^{(0)} = \mathbf{\Phimat}^\top\yv$; $f_\theta(\cdot\ ; \yv, \Phimat)$ is an iteration map ${\mathbb R}^{nB} \rightarrow {\mathbb R}^{nB}$ towards a stable equilibrium:
\begin{align}
    \lim_{k \rightarrow + \infty} \xv^{(k)} &= \lim_{k \rightarrow + \infty} f_\theta (\xv^{(k)}; \yv, \Phimat) \nonumber\\
    &\equiv \hat{\xv} = f_\theta (\hat{\xv}; \yv, \Phimat),  \label{eq:eq}
\end{align}
where $\hat{\xv}\in {\mathbb R}^{nB}$ denotes the fixed point and reconstruction result.

In following sections, we design different $f_\theta$ for SCI, in terms of the implicit infinite-depth RNN architecture and infinitely iterative PnP framework. Following~\cite{zhao2022deep,gilton2021deep}, we utilize Anderson acceleration~\cite{walker2011anderson} to compute the fixed point of $f_\theta$ efficiently in Sec.~\ref{sec:forward}. Following~\cite{zhao2022deep,gilton2021deep}, for gradient calculation, we optimize the network wights $\theta$ by approximating the inverse Jacobian, described in Sec.~\ref{sec:backward}. Convergence of this scheme for specific $f_\theta$ designs is discussed in Sec.~\ref{sec:convergence}.

% \section{Joint Framework}
% \subsection{Deep Equilibrium Gradient Descent}

% \subsection{Deep Equilibrium Proximal Gradient Descent}

% \subsection{Deep Equilibrium ADMM}

\section{Forward Pass}
\label{sec:forward}
Unlike the conventional optimization method where the terminal step number is manually chosen or a network where the output is the activation from the limited layers, the result of DEQ is the equilibrium point itself. Therefore, the forward evaluation could be any procedure
that solves for this equilibrium point. Considering SCI reconstruction, we design novel iterative models that converge to equilibrium.

\subsection{Recurrent Neural Networks}

\begin{figure}
    \centering
    \includegraphics[width=0.7\linewidth]{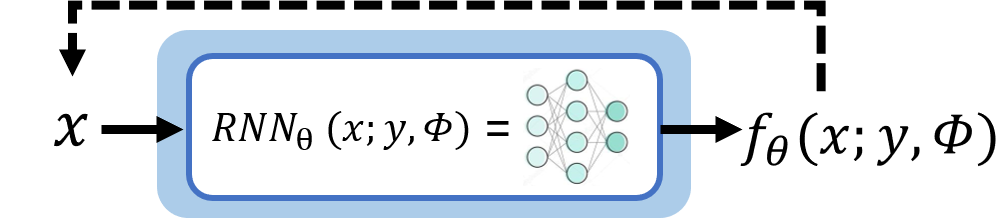}
    \caption{Illustration of our proposed DEQ for SCI using recurrent neural network (RNN), \ie, DE-RNN.}
    \label{fig:pipeline_rnn}
\end{figure}

To achieve integration of DEQ and RNN for video SCI, we have:
\begin{align}
    \xv^{(k+1)} = {\rm RNN}_\theta (\xv^{(k)}, \yv, \Phimat),
\label{eq:dernn-x}
\end{align}
where ${\rm RNN}(\cdot\ )$ is a trainable RNN network learning to iteratively reconstruct effective and stable data. As shown in Fig~\ref{fig:pipeline_rnn}, the corresponding iteration map is:
%%%%%%%%%%
\begin{align}
    f_\theta (\xv; \yv, \Phimat) = {\rm RNN}_\theta (\xv, \yv, \Phimat).
\label{eq:dernn-f}
\end{align}

\subsection{Generalized Alternating Projection}

\begin{figure}
    \centering
    \includegraphics[width=0.9\linewidth]{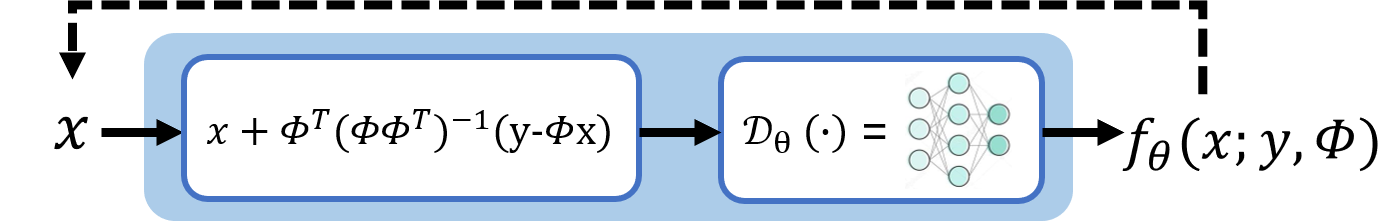}
    \caption{Illustration of our proposed DEQ for SCI using generalized alternating projection (GAP), \ie, DE-GAP.}
    \label{fig:pipeline_degap}
\end{figure}

Regarding the optimization iterations in the GAP method, represented in Eq.~\eqref{eq:gap-x}-\eqref{eq:gap-v}, we iteratively update $\xv$ by: 
%%%%%%%%%%%%
\begin{align}
    \xv^{(k+1)} = \textstyle \mathcal{D}^{(k+1)}_\theta \left[\xv^{(k)} + {\Phimat}^\top (\Phimat {\Phimat}^\top)^{-1}(\yv-\Phimat \xv^{(k)}) \right].
\label{eq:degap-x}
\end{align}
%%%%%%%%%%%%%%%%%%
Therefore, as illustrated in Fig.~\ref{fig:pipeline_degap}, the iteration map is:
%%%%%%%%%%%%%%%%%
\begin{align}
    f_\theta (\xv; \yv, \Phimat) = \textstyle \mathcal{D}_\theta(\xv + {\Phimat}^\top (\Phimat {\Phimat}^\top)^{-1}(\yv-\Phimat \xv) ).
\label{eq:degap-f}
\end{align}

\subsection{Anderson Acceleration}
To enforce fixed-point iterations converge more quickly, we make full use of the ability to accelerate inference with standard fixed-point accelerators, \textit{e.g.}, Anderson accelerator. Anderson acceleration utilizes previous iterations to seek promising directions to move forward. Under the setting of Anderson accelerator, we identify a vector $\alphav^{(k)}\in\mathbb{R}^s$, for $\delta > 0$:
%%%%%%%%%%%%%%%%%%
\begin{align}
\begin{aligned}
    \xv^{(k+1)} = & \textstyle (1-\delta) \sum_{i=0}^{s-1} \mathbf{\alphav}_i^{(k)} \xv^{(k-i)} \\
    &+ \textstyle \delta \sum_{i=0}^{s-1} \mathbf{\alphav}_i^{(k)} f_\theta (\xv^{(k-i)}; \yv, \Phimat), 
\end{aligned}
\end{align}
%%%%%%%%%%%%%
where the vector $\mathbf{\alphav}_i^{(k)}$ is the solution to the optimization problem:
%%%%%%%%%%%%%%%
\begin{align}
  \textstyle  \argmin_\mathbf{\alphav} || \Amat \alphav ||_2^2,\quad s.t.\quad \mathbf{1}^\top\alphav = 1,
    \label{eq:anderson}
\end{align}
%%%%%%%%%%%
where $\Amat$ is a matrix whose $i$-{th} column is the vectorized residual $f_\theta (\xv^{(k-i)}; \yv, \Phimat) - \xv^{(k-i)}$, with $i=0,\dots,s-1$. When $s$ is small (\textit{e.g.}, $s = 3$), the optimization problem in Eq.~\eqref{eq:anderson} introduces trivial computation.

\section{Backward Pass}
\label{sec:backward}
While previous work often utilizes Newton’s method to achieve the equilibrium and then backpropagate through all the Newton iterations, following~\cite{zhao2022deep,gilton2021deep}, we alternatively adopt another method with high efficiency and constant memory requirement.

% Using DEQ, we no longer rely on explicit backpropagation through the exact operations in the forward pass. While one can certainly fix an algorithm (\textit{e.g.}, Newton’s method) to obtain the equilibrium, and then store and backpropagate through all the Newton iterations, we provide below an alternative procedure that is much more efficient, and requires only constant memory.
%%%%%%%%%%%%%
\subsection{Loss Function}
To optimize network parameters $\theta$, stochastic gradient descent is used to minimize a loss function as follows:
\begin{align}
    % \argmin_\theta = \lim_{k \rightarrow + \infty} \frac{1}{n} \sum_{i=1}^{n} \it{l}(f_\theta (\xv^{(k)}_i; \yv_i, \Phimat_i), \xv^\star_i),
   \textstyle  \theta^* =   \argmin_\theta \frac{1}{m} \sum_{i=1}^{m} \ell (f_\theta (\hat{\xv}_i; \yv_i, \Phimat_i), \xv^\star_i),  
\end{align}
where $m$ is the number of training samples; $\ell(\cdot, \cdot)$ is a given loss function, $\xv^\star_i$ is the ground truth 3D data of the $i$-{th} training sample, $\yv_i$ is the paired measurement, $\Phimat_i$ denotes the sensing matrix, and $f_\theta (\hat{\xv}_i; \yv_i, \Phimat_i)$ denotes the reconstruction result given as the fixed point $\hat{\xv}$ of the iteration map $f_\theta (\cdot\ ; \yv, \Phimat)$, as derived from Eq.~\eqref{eq:eq}.
%%%%%%%%%%%%%%%%%%%%
The mean-squared error (MSE) loss is used for our video SCI reconstruction:
\begin{align}
    \textstyle  \ell(\hat{\xv}, \xv^\star) = \frac{1}{2} || \hat{\xv} - \xv^\star ||_2^2.
\label{eq:mseloss}
\end{align}
%%%%%%%%%%%%%
Since the reconstruction result is a fixed point of the iteration map $f_\theta (\cdot\ ; \yv, \Phimat)$, gradient calculation of this loss term could be designed to avoid large memory demand. Following~\cite{gilton2021deep}, we calculate the gradient of the loss term, which takes the network parameters $\theta$ into consideration. 
% with backpropagating through an arbitrarily-large number of fixed-point iterations.

\subsection{Gradient Calculation}
Following~\cite{zhao2022deep,gilton2021deep}, we calculate the loss gradient. Let $\ell$ be an abbreviation of $\ell(\hat{\xv}, \xv^\star)$ in Eq.~\eqref{eq:mseloss}, then the loss gradient is: 
%%%%%%%%%%%%%%%%%%%%%
\begin{align}
\begin{aligned}
\label{eq:gradient}
    \textstyle \frac{\partial \ell}{\partial \theta} &= {\left(\frac{\partial \hat{\xv}}{\partial \theta}\right)}^\top \frac{\partial \ell}{ \partial \hat{\xv}}
    &= {\left(\frac{\partial \hat{\xv}}{\partial \theta}\right)}^\top {\left(\hat{\xv} - \xv^\star \right)},
\end{aligned}
\end{align}
%%%%%%%%%%
where $\frac{\partial \hat{\xv}}{\partial \theta}$ is the Jacobian of $\hat{\xv}$ evaluated at $\theta$, and $\frac{\partial \ell}{ \partial \hat{\xv}}$ is the gradient of $\ell$ evaluated at $\xv^\star$.

Then to compute the Jacobian $\frac{\partial \hat{\xv}}{\partial \theta}$, we recall the fixed point equation $\hat{\xv} = f_\theta (\hat{\xv}; \yv, \Phimat)$ in Eq.~\eqref{eq:eq}. By implicitly differentiating both sides of this fixed point equation, the Jacobian $\frac{\partial \hat{\xv}}{\partial \theta}$ is solved as:
\begin{align}
   \textstyle  \frac{\partial \hat{\xv}}{\partial \theta} = \left[\boldsymbol{I} - \left.\frac{\partial f_\theta (\xv; \yv, \Phimat)}{\partial \xv}\right|_{\xv = \hat{\xv}}\right]^{-1} \frac{\partial f_\theta (\hat{\xv}; \yv, \Phimat)}{\partial \theta},
\end{align}
which could be plugged into Eq.~\eqref{eq:gradient} and thus get:
\begin{align}
\begin{aligned}
    \textstyle \frac{\partial \ell}{\partial \theta} =   \left[\frac{\partial f_\theta (\hat{\xv}; \yv, \Phimat)}{\partial \theta}\right]^\top \! {\left[\boldsymbol{I} - \left.\frac{\partial f_\theta (\xv; \yv, \Phimat)}{\partial \xv}\right|_{\xv = \hat{\xv}}\right]^{-\top}} 
     {(\hat{\xv} - \xv^\star)},
    \end{aligned}
\end{align}
where $^{-\top}$ denotes the inversion followed by transpose.  
As this method converted gradient calculation to the problem of calculating an inverse Jacobian-vector product, it avoids the backpropagation through many iterations of $f_\theta (\hat{\xv}; \yv, \Phimat)$.  To approximate the inverse Jacobian-vector product, we define the vector $\av^{(\infty)}$ as:
\begin{align}
     \textstyle \av^{(\infty)} = {\left[\boldsymbol{I} - \left.\frac{\partial f_\theta (\xv; \yv, \Phimat)}{\partial \xv}\right|_{\xv = \hat{\xv}}\right]^{-\top}} \! {(\hat{\xv} - \xv^\star)}.
\end{align}
Following~\cite{gilton2021deep}, it is noted that $\av^{(\infty)}$ is a fixed point of the equation:
%%%%%%%%%%%%%%%%%
\begin{align}
\begin{aligned}
\label{eq:jac-eq}
    \av^{(k+1)} =    \textstyle {\left[\left.\frac{\partial f_\theta (\xv; \yv, \Phimat)}{\partial \xv}\right|_{\xv = \hat{\xv}}\right]^{-\top}}\av^{(k)} + {(\hat{\xv} - \xv^\star)},\\ \forall k = 0,1,\dots, \infty.
\end{aligned}
\end{align}
Therefore, the same algorithm used to calculate the fixed point $\hat{\xv}$ could also be used to calculate $\av^{(\infty)}$ . The limit of fixed-point iterations for solving Eq.~\eqref{eq:jac-eq} with initial iterate $\av^{(0)} = \mathbf{0}$ is denoted equivalently to the Neumann series:
\begin{align}
\label{eq:neumann}
    \av^{(\infty)} =  \textstyle \sum_{p=0}^{\infty} 
    \left\{
    {\left[\left.\frac{\partial f_\theta (\xv; \yv, \Phimat)}{\partial \xv}\right|_{\xv = \hat{\xv}}\right]^{\top}}
    \right\}
    ^{p}\!
    {(\hat{\xv} - \xv^\star)}.
\end{align}

To quickly calculate the vector-Jacobian products in Eq.~\eqref{eq:jac-eq} and
Eq.~\eqref{eq:neumann}, a lot of auto-differentiation tools (\textit{e.g.}, autograd packages in Pytorch\cite{paszke2019pytorch}) could be utilized. After the accurate approximation of $\av^{(\infty)}$ is calculated, the gradient in Eq.~\eqref{eq:gradient} is given by:
%%%%%%%%%%%%%%%%%%%%%%%%%%55
\begin{align}
     \textstyle\frac{\partial \ell}{\partial \theta} = {\left(\frac{\partial f_\theta (\hat{\xv}; \yv, \Phimat)}{\partial \theta}\right)}^\top \av^{(\infty)}.
\end{align}

\section{Convergence Theory}
\label{sec:convergence}
Given the iteration map $f_\theta(\cdot\ ; \yv, \Phimat): {\mathbb R}^{nB} \rightarrow {\mathbb R}^{nB}$, in this section, we discuss conditions that guarantee the convergence of the proposed deep equilibrium models $\xv^{(k+1)} = f_\theta (\xv^{(k)}; \yv, \Phimat)$ to a fixed-point $\hat{\xv}$ as $k\rightarrow\infty$.   

%\xin{Xin: I AM HERE!!!}

%%%%%%%%%%%%%
\begin{assumption}
\textbf{(Convergence of DE-RNN)}.
For all $\xv, \xv' \in {\mathbb R}^{nB}$, if there exists a constant $0\leq c <1$ satisfies that:
\begin{align}
    \| {\rm RNN}_\theta (\xv, \yv, \Phimat) -  {\rm RNN}_\theta (\xv', \yv, \Phimat)\|\leq c\| \xv - \xv' \|,
\label{eq:dernn-conver}
\end{align}
then the DE-RNN iteration map $f_\theta (\xv; \yv, \Phimat)$ is contractive.
\end{assumption}

\begin{assumption}
\textbf{(Convergence of DE-GAP)}.
For all $\xv, \xv' \in {\mathbb R}^{nB}$, if there exists a $\varepsilon > 0$ such that the denoiser $\mathcal{D_\theta}: {\mathbb R}^{nB}\rightarrow{\mathbb R}^{nB}$ satisfies:
\begin{align}
\label{eq:assum}
    \|(\mathcal{D}_\theta - \Imat)(\xv) - (\mathcal{D}_\theta - \Imat)(\xv') \| \leq \varepsilon || \xv - \xv'||,
\end{align}
where $ (\mathcal{D}_\theta - \Imat)(\xv) := \mathcal{D}_\theta(\xv) - \xv$, that is, we assume the map $\mathcal{D}_\theta - \Imat$ is $\varepsilon$-Lipschitz, 
then the DE-GAP iteration map $ f_\theta (\cdot ; \yv, \Phimat) $ defined in Eq.~\eqref{eq:degap-f} satisfies:
\begin{align}
 \textstyle    \| f_\theta (\xv ; \yv, \Phimat) - f_\theta (\xv' ; \yv, \Phimat)\| \leq \eta \| \xv - \xv' \|
\label{eq:degap-conver}
\end{align}
for all $\xv, \xv' \in {\mathbb R}^{nB}$. The coefficient $\eta$ is less than 1, in which case the DE-GAP iteration map $f_\theta (\xv; \yv, \Phimat)$ is contractive.
% , where $\eta = (1+\varepsilon) \max_i| 1-\lambda_i |$, and $\lambda_i$ are eigenvalues of ${\Phimat}^\top(\Phimat {\Phimat}^\top)^{-1}\Phimat$
\end{assumption}
% \begin{proof}
% \textit{Proof.} 
% Following ~\cite{ryu2019plug}, we assume that 

Following~\cite{gilton2021deep}, to prove $f_\theta(\cdot ; \yv, \Phimat)$ is contractive it suffices to show $|| \partial_{\xv} f_\theta(\xv; \yv, \Phimat)|| < 1 $ for all $\xv \in {\mathbb R}^{nB}$, where $|| \cdot ||$ denotes the spectral norm, $\partial_{\xv} f_\theta(\xv; \yv, \Phimat)$ is the Jacobian of $f_\theta(\xv; \yv, \Phimat)$ with respect to $\xv \in {\mathbb R}^{nB}$ given by:
\begin{align}
%\begin{aligned}
    \partial_{\xv} f_\theta(\xv; \yv, \Phimat) &= \partial_{\xv} \mathcal{D_\theta}(\xv) (\Imat - {\Phimat}^\top (\Phimat {\Phimat}^\top)^{-1}\Phimat),
%\end{aligned}
\end{align}
where $\partial_{\xv} \mathcal{D_\theta} \in {\mathbb R}^{nB\times nB}$ is the Jacobian of $\mathcal{D_\theta}: {\mathbb R}^{nB}\rightarrow{\mathbb R}^{nB}$ with respect to $\xv \in {\mathbb R}^{nB}$.  

Finally, we derive (details can be found in~\cite{zhao2022mathematical} or supplementary material):
%%%%%%%%%%%%%%%%%%%%%%%%
%%%%%%%%%%%%%%%%%%%%%%%%
%%%%%%%%%%%%%%%%%%%%%%%%
\begin{align}
\begin{aligned}
    || \partial_{\xv} f_\theta(\xv; \yv, \Phimat) ||
    = &\ || \partial_{\xv} \mathcal{D_\theta}(\xv) (\Imat - \Psimat) ||, \\
    = &\ ||\partial_{\xv} \mathcal{D_\theta}(\xv) -  \partial_{\xv} \mathcal{D_\theta}(\xv) \Psimat || \\
    = &\ ||\partial_{\xv} \mathcal{D_\theta}(\xv) - \Imat + \Imat -  \partial_{\xv} \mathcal{D_\theta}(\xv) \Psimat || \\
    \leq &\ ||\partial_{\xv} \mathcal{D_\theta}(\xv) - \Imat || + || \Imat -  \partial_{\xv} \mathcal{D_\theta}(\xv) \Psimat || \\
    \leq &\ \varepsilon + 1 + || (\partial_{\xv} \mathcal{D_\theta}(\xv) - \Imat + \Imat) \Psimat || \\
    = &\ 1 + \varepsilon + || (\partial_{\xv} \mathcal{D_\theta}(\xv) - \Imat) \Psimat + \Psimat || \\
    \leq &\ 1 + \varepsilon + || (\partial_{\xv} \mathcal{D_\theta}(\xv) - \Imat) \Psimat ||  + || \Psimat || \\
    \leq &\ 1 + \varepsilon + \varepsilon || \Psimat || + || \Psimat || \\
    = &\ (1 +\varepsilon) (1 + || \Psimat ||) \\
     % \leq \eta,
    \leq &\textstyle (1+\varepsilon) \max_i| 1-\lambda_i |, 
    \label{eq:inequ}
\end{aligned}
\end{align}
where $\lambda_i$ are eigenvalues of ${\Phimat}^\top (\Phimat {\Phimat}^\top)^{-1}\Phimat $; and the inequality Eq.~\eqref{eq:inequ} is based on the assumption that the map $(\mathcal{D}_\theta - \Imat)(\xv) := \mathcal{D}_\theta(\xv) - \xv$ is $\varepsilon$-Lipschitz.
Therefore the spectral norm of its Jacobian $\partial_{\xv} \mathcal{D_\theta}(\xv) - \Imat$ is bounded by $\eta$, which demonstrates $f_\theta$ is $\eta$-Lipschitz with $\eta = (1+\varepsilon) \max_i| 1-\lambda_i |$. 

It is worth noting that convergence is not yet guaranteed in our calculation above since $\eta$ is larger than 1. In SCI cases, it is challenging to provide a theoretical guarantee. However, we observe our models converge well in the experiments.